\documentclass[10pt,twocolumn,letterpaper]{article}

\usepackage{iccv}
\usepackage{times}
\usepackage{epsfig}
\usepackage{graphicx}
\usepackage{amsmath}
\usepackage{amssymb}
\usepackage{multirow}
\usepackage{subfigure}
\usepackage{caption}
\usepackage[pagebackref=true,breaklinks=true,letterpaper=true,colorlinks,bookmarks=false]{hyperref}

\iccvfinalcopy % *** Uncomment this line for the final submission

\def\iccvPaperID{1} % *** Enter the ICCV Paper ID here

\ificcvfinal\fi

\begin{document}

%%%%%%%%% TITLE
\title{STRIDE: Street View-based Environmental Feature Detection and Pedestrian Collision Prediction}

% Universidad de los Andes\\

\author{Cristina Gonz\'{a}lez$^{1,2}$*$
\qquad$Nicol\'{a}s Ayobi$^{1,2}$*$
\qquad$Felipe Escall\'{o}n$^{1,2}$\\ %$
Laura Baldovino-Chiquillo$^3$$
\qquad$Maria Wilches-Mogoll\'{o}n$^2$$
\qquad$Donny Pasos$^4$$
\qquad$Nicole Ram\'{i}rez$^2$\\ %
Jose Pinz\'{o}n$^{3,6}$$
\qquad$Olga Sarmiento$^3$$
\qquad$D. Alex Quistberg$^{3,5}$$
\qquad$Pablo Arbel\'{a}ez$^{1,2}$\\
\small{$^1$Center for Research and Formation in Artificial Intelligence, Universidad de los Andes, Colombia}\\
\small{$^2$School of Engineering, Universidad de los Andes, Colombia}
\quad \small{$^3$School of Medicine, Universidad de los Andes, Colombia}\\
\small{$^4$School of Economics, Universidad de los Andes, Colombia}
\quad \small{$^5$Dornsife School of Public Health, Drexel University, USA}\\
\small{$^6$School of Architecture and Design, Pontificia Universidad Javeriana, Colombia}\\[2mm]
}

% \and

% \author{Cristina González* \and Nicolás Ayobi* \and Felipe Escallon \and Laura Baldovino \and Maria Wilches-Mogollon \and Donny Pasos \and Nicole Ramírez \and José Pinzón \and Olga Sarmiento \and Alex Quistberg \and Pablo Arbeláez\\
% }

\maketitle
\ificcvfinal\thispagestyle{empty}\fi

\newcommand\blfootnote[1]{%
  \begingroup
  \renewcommand\thefootnote{}\footnote{#1}%
  \addtocounter{footnote}{-1}%
  \endgroup
}

%%%%%%%%% ABSTRACT
\begin{abstract}

This paper introduces a novel benchmark to study the impact and relationship of built environment elements on pedestrian collision prediction, intending to enhance environmental awareness in autonomous driving systems to prevent pedestrian injuries actively. We introduce a built environment detection task in large-scale panoramic images and a detection-based pedestrian collision frequency prediction task. We propose a baseline method that incorporates a collision prediction module into a state-of-the-art detection model to tackle both tasks simultaneously. Our experiments demonstrate a significant correlation between object detection of built environment elements and pedestrian collision frequency prediction. Our results are a stepping stone towards understanding the interdependencies between built environment conditions and pedestrian safety.

\end{abstract}
\blfootnote{*Equal contribution}

%%%%%%%%% BODY TEXT
% ---------------- INTRODUCTION ----------------------
\section{Introduction}\label{sec:introduction}
Autonomous driving systems rely on their ability to gather and interpret information from their surroundings~\cite{niemeijer2017review}, enabling them to anticipate future events and make situation-aware decisions without compromising road safety for all parties involved. A challenging task for autonomous vehicles (AVs) is the detection of pedestrians and other vulnerable road users to avoid pedestrian-motor vehicle collisions. While much of the prior research has focused on pedestrian-detection tasks, few studies have examined the role of road infrastructure and built environment features in improving pedestrian detection and thus reducing the chance of pedestrian collisions. In particular, it is well known that pedestrian safety is impacted by road design and the built environment~\cite{sisiopiku2003pedestrian,quistberg2015walking,craig2019pedestrian,foster2014evaluating,donroe2008pedestrian,retting2003review,want_to_cross}, including objects that comprise defined pedestrian crossing areas (e.g., crosswalks, stop lines, speed bumps), traffic control (e.g., traffic signs, pedestrian signs, stop signs) and traffic speed (e.g., road width, traffic lanes, trees, street lights). 

\begin{figure}[]
    \centering
    \includegraphics[width=\linewidth]{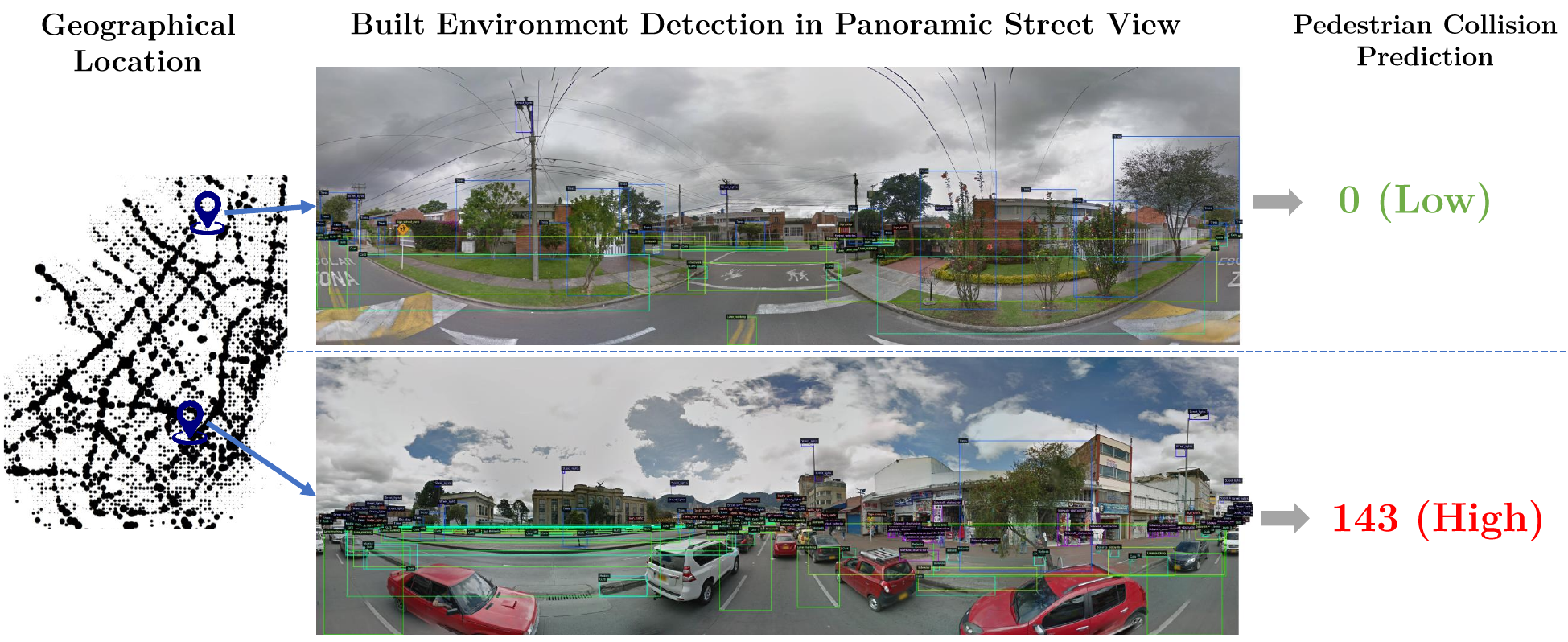}
    \caption{\textbf{STRIDE.} Given specific city coordinates (left) and the corresponding panoramic street view image (middle), we propose to predict the number of pedestrian collisions in those coordinates (right) by detecting built environment features (middle).}
    \label{fig:examples}
\end{figure}

However, the influence of these features on pedestrian injuries can vary depending on the dynamics of road users within specific geographical locations. Consequently, the presence or absence of such features may have distinct effects in low and middle-income countries compared to high-income ones~\cite{quistberg2015walking}. Therefore, it is crucial to prioritize advancing autonomous driving systems that can generalize to the unique characteristics of their environments worldwide.

Despite road features' crucial role in pedestrian injuries, limited research has explored the correlation between these objects and pedestrian collision frequency using visual information extracted from the street-level scene. Existing frameworks often rely on precomputed data about the street environment or primarily focus on identifying and anticipating collisions rather than proactively preventing them. This research gap underscores the need for a comprehensive approach that leverages visual cues from the street to analyze the relationship between road features and pedestrian collision occurrences. 

This paper introduces a novel dataset comprising more than 18k Google Street View~\cite{anguelov2010google} panoramic images, annotated with bounding boxes for 27 categories of common road environment objects that may affect pedestrian safety. Furthermore, we calculate the true incidence of pedestrian injuries for specific geographical points corresponding to the images within the dataset by leveraging public historical records from 2015 to 2021 from Bogota City in Colombia.

Most publicly available data for autonomous driving primarily originates from European, Asian, or North American countries~\cite{Cityscapes, mapillaryvistas, bdd100k, kitty, us_accidents}, thus leaving Latin American and African countries significantly underrepresented. Therefore, focusing on the Latin American region bridges a crucial gap and provides specific insights into these countries and their unique challenges. The geographic diversity introduced by our dataset facilitates the development of more inclusive and robust models for autonomous driving and other related applications.

Thus, our approach addresses both the object detection task and predicts the frequency of pedestrian collisions. We introduce a baseline method that builds upon the state-of-the-art model DINO~\cite{zhang2022dino}. Our model can estimate the number of pedestrian collisions associated with a given location by leveraging actual visual features from the images and the corresponding geographical coordinates.

Our main contributions can be summarized as follows:
\begin{enumerate}
    \item We propose the task of automated pedestrian collision prediction by considering the road-built environment in a specific location. 
    \item We establish an experimental framework for studying this problem in a city within the Latin American context, including frequencies for pedestrian collisions and detection labels for Google Street View panoramic images.
    \item We empirically demonstrate that by training a multi-task model to detect objects in the urban built environment with a potential influence on pedestrian injuries, we can improve the predicting capabilities of the model.
\end{enumerate}

To ensure the reproducibility of our results and to promote further research on predicting pedestrian collisions, we make all the resources of this paper publicly available on our project web page \footnote{\url{https://github.com/BCV-Uniandes/STRIDE}}.
% our benchmark dataset annotations, the pretrained models, and the source code for our baseline.

% ------------------ RELATED WORK -----------------------------
\section{Related Work}\label{sec:related-work}
\begin{figure*}[ht!]
    \centering
    \includegraphics[width=\textwidth]{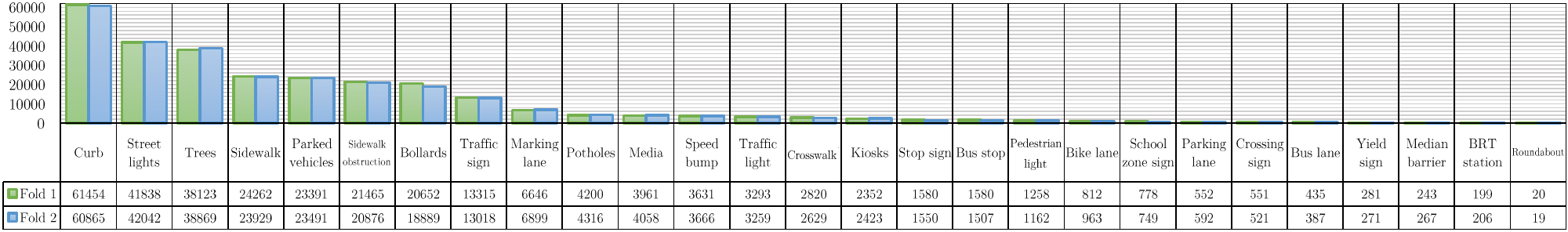}
    \caption{Number of labeled bounding boxes (y-axis) per class in each fold and their corresponding category name (x-axis).}
    \label{fig:classes_distr}
\end{figure*}
\subsection{Built Environment Object Recognition in Autonomous Driving}

Object recognition for autonomous driving has been extensively studied. For instance, generic object detection benchmarks like MS COCO~\cite{mscoco}, and PascalVOC~\cite{pascalvoc} include street images with annotated vehicles, pedestrians, and some general road elements. More specifically, pedestrian detection is a pioneering and well-studied task~\cite{hogsvm, eth_pedestrian, Caltech_pedestrian}, explored with large and complex datasets ~\cite{CityPersons, eurocitypersons, nightowls, kaist, tsinghua}, and even with 3D detection and tracking approaches~\cite{person_tracking, kitty}. Broader benchmarks also cover detection of vehicles~\cite{RoViT, d_city, tme}, traffic signs~\cite{RoViT, pano_rsod, mapillary_traffic}, and axis-aligned vehicle detection~\cite{boxy, tme}. However, these datasets focus mainly on dynamic agents and neglect the importance of static road infrastructure elements for comprehensive scene understanding.

% Object recognition in the built environment context has been extensively studied for many years and multiple benchmarks have been developed to address the specific challenges autonomous driving. For instance, generic object recognition datasets like MS COCO \cite{mscoco} and PascalVOC \cite{pascalvoc} include multiple street scene images with vehicles, pedestrians, and some road elements annotated. However, one of the field's pioneering and most studied tasks is pedestrian detection \cite{hogsvm, eth_pedestrian, Caltech_pedestrian}. This task has nowadays been extensively investigated with large-scale datasets \cite{CityPersons, eurocitypersons}, high variability scenes \cite{nightowls, kaist}, inclusion of riders \cite{nightowls, tsinghua}, and 3D detection and tracking \cite{person_tracking, kitty}. Alternatively, broader detection benchmarks also include vehicles detection \cite{RoViT, d_city}, traffic signs detection\cite{RoViT, pano_rsod, mapillary_traffic}, vehicle tracking \cite{d_city, tme}, and axis aligned vehicle detection \cite{boxy, tme}. Nevertheless, most of these datasets focus on studying dynamic objects within urban environments, often overlooking the significance of static road infrastructure elements contributing to comprehensive scene understanding.

On the contrary, standard benchmarks for urban static object identification include lane segmentation~\cite{culane, vpgnet}, lane markings detection~\cite{vpgnet, pano_rsod}, and multiple traffic signs detection~\cite{mapillary_traffic}. Fine-grained urban scene parsing datasets like~\cite{camvid,dus,mit_driveseg} provide pixel-level semantic annotations for street images, covering some static road infrastructure and dynamic agents. CityScapes~\cite{Cityscapes} and KITTY~\cite{kitty_segmentation} offer panoptic segmentation annotations, thus identifying instances within some semantic classes. Mapillary Vistas \cite{mapillaryvistas} extended this framework by including many more semantic categories and using highly variable user-uploaded data. BDD100K~\cite{bdd100k} extends these tasks through time by including segment tracking in videos, while ApolloScape~\cite{apolloscape}, KITTI~\cite{kitty, semantic_kitty, semantic_kitty2} and \cite{semantic_2d_3d}, focus on 3D point cloud segmentation. However, most of these benchmarks have limited classes for street infrastructure, often treating them as stuff categories, unlike our benchmark, which explicitly identifies and differentiates these objects to study their effect on pedestrian collision frequency.

\subsection{Panoramic Street View Benchmarks} 

Initial frameworks for wide field of view autonomous driving tackle object detection of a few categories using panoramic images~\cite{pano_rsod, panoramic_detections, multi-view}. Further approaches use data gathered with fisheye and surround-view cameras annotated for semantic scene parsing~\cite{RestrictedDC, woodscape, near-field, OmniDetSV, Kitti360}, thus introducing significant distortion and deformation caused by this type of cameras. Other frameworks use synthetic data~\cite{OmniScape, synthia-pano} for the same task, which generates a considerable gap for real-world applications. 

Fine-grained frameworks offer pixel-wise annotations on a few panoramas from annular lenses~\cite{pass, DSPASSDP} or Google Street Views~\cite{wildpass, wildpass_2, wildpps, p2pda, CVRG-Pano} as testing sets for domain adaptation. Recently, Mapillary Metropolis~\cite{MapillaryMetropolis} introduced the first large-scale panoramic panoptic segmentation benchmark with 360° panoramas of 4,000$\times$8,000 resolution aligned with aerial images and 3D point clouds. Similarly, the Waymo Panoramic Video Panoptic Segmentation Dataset~\cite{waymo_panoramic} used an extensive set of 220° panoramas for panoptic segmentation and segment tracking. Despite these advancements, many street infrastructure classes are still disregarded in these benchmarks, and only Mapillary Metropolis utilizes large panoramic images. On the contrary, our dataset considers multiple streets features categories on 4,000$\times$13,312 images, surpassing any previous panoramic benchmark in image size.

\subsection{Street Collision Prediction Benchmarks}

Existing methods for collision frequency prediction rely primarily on tabular variables related to road properties, street conditions, traffic volume, and environmental factors~\cite{data_mining,artificial_networks,carr_crash, TAP-CNN, ta_stan}. More complex approaches incorporate historical collision records~\cite{sdcae,Time_coords}, spatial and temporal relations among city regions and time windows~\cite{spatio_temporal, spatio_temporal_2,spatio_temporal_4, hetero_convlstm}, and satellite images~\cite{satellite, hetero_convlstm}. Conversely, the US-Accidents benchmark~\cite{us_accidents,us_accidents_2} provides information on the presence of general street components. However, these frameworks rely heavily on alternative or precalculated data rather than analyzing urban scene images captured from street-level perspectives.

In contrast, computer vision-based approaches predominantly employ detection and tracking methods to identify vehicle crashes \cite{xu2022tad, yao2020when, unsupervised_accidents_detection, deep_learning_detection_local, stacked_autoencoders, DETR_accidents, Mask-RCNN_accidents} or to track vehicles and other agents to predict accidents and anomalies \cite{4th_ai_city_challenge, 5th_ai_city_challenge, dota, vision_based_anomaly_detection, CADP, unsupervised_first-person, real_time_accident_detection, real_time_framework, Zhao2019UnsupervisedTA, multi_granularity}. Similarly, some frameworks estimate vehicle trajectories to anticipate collisions \cite{CollJosifov2022, collide_pred, intelligent_intersection}. Other alternative benchmarks utilize reinforcement learning \cite{reinforcement}, forecasting in time series \cite{kosman2021visionguided}, and causality recognition \cite{causality} to anticipate accidents. Contrarily, some datasets specifically target pedestrian safety by predicting pedestrians' intentions to cross \cite{pie,goint_to_cross}, and some studies on this task have demonstrated that street infrastructure state considerably impacts pedestrian crossing prediction capabilities~\cite{want_to_cross}. However, no previous framework has directly studied the relations between built environment elements and collision prediction using computer vision models.

\section{STRIDE Dataset}\label{sec:dataset}
\begin{figure}[]
    \centering
    \includegraphics[width=\linewidth]{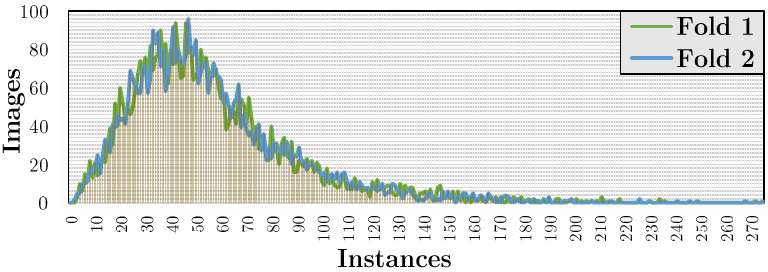}
    \caption{\textbf{Distribution of the number of bounding box annotations per image.} The figure shows the number of images (y-axis) with a certain number of annotated instances (x-axis). Our images contain a varying range of instances with a similar distribution among folds.}
    \label{fig:annotations_distr}
\end{figure}

We introduce the Street View-based Environmental Feature and Pedestrian Collision Dataset (STRIDE), a novel challenging benchmark that studies the interrelations between built-environment elements and pedestrian collision frequency for scene awareness in autonomous driving. Figure \ref{fig:examples} presents an overview of our benchmark. STRIDE combines multiple public data sources for two main tasks: (1) road-built-environment static object detection; and (2) image-guided pedestrian collision frequency estimation. In this section, we describe the details of our benchmark.

\subsection{Image Gathering}

First, we uniformly sampled random locations along the streets of Bogota City in Colombia, including specific points where statistical analysis indicated a higher incidence of pedestrian injuries. Secondly, to leverage complete 360° information, we utilize the 3D Google Street View service~\cite{StreetView} to download panoramic images corresponding to the selected locations. The resulting dataset encompasses 18,036 panoramic images from different parts of the city. Our images have a high resolution of 4,000$\times$13,312, making them the panoramic dataset with the most extensive images. We split our data into a training and validation set and a test set. Our training and validation set comprises 9,900 images on which we performed a 2-fold cross-validation to train and validate our experiments. The remaining 8,136 images are used for testing.

\subsection{Detection Annotations}

\begin{figure}[]
     \centering
     \includegraphics[width=0.95\linewidth]{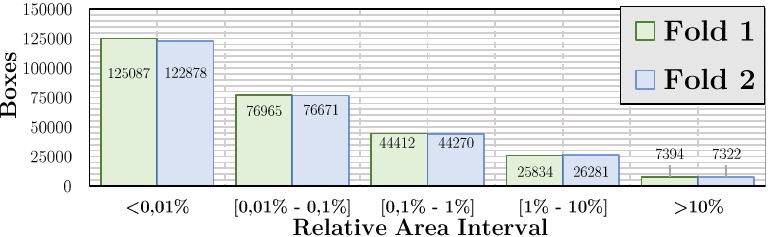}
     \caption{\textbf{Number of annotated bounding boxes (y-axis) per relative box area interval (x-axis).} Most instances in our dataset have small relative sizes due to the large scale of our images.}
     \label{fig:rel_area}
 \end{figure}

We selected a team of skilled annotators and trained them to draw bounding boxes around relevant static objects in the street infrastructure present in our panoramic images. Experts in built environment safety defined the classes used for annotation to encompass a comprehensive range of static objects that play crucial roles in pedestrian safety. We annotated all 9,900 images in our training set with 27 static street furniture classes indicated in Figure \ref{fig:classes_distr}. For cross-validation splitting, we ensured that the distributions of street object classes, number of boxes per image, and area of boxes were maintained consistently across both folds. 

% \subsection{Detection Annotation Consistency}

% To assess the consistency of the annotations, we randomly selected 992 duplicate images, which were annotated separately.

\subsection{Pedestrian Collision Annotations}

To correlate our dataset with real-world pedestrian collisions data, we leverage publicly available pedestrian collision records of Bogota City from 2015 until 2021. This data consists of records documenting the total number of vehicle collisions with pedestrians (pedestrian collisions) reported for each crossing point on the city's streets. By utilizing the corresponding geographic coordinates of each image, we associate each of our panoramic images with a single crossing point. We used ArcGIS 10.8 to perform geospatial analysis and geostatistics to identify each image's closest registered crossing point. Most of our images matched a registered point within a 30-meter distance; the rest were matched to points in a 100-meter radius. We filtered a few images outside of the urban area or with more than a 100-meter distance to a crossing point. We provide the distribution of distances in the Supplementary Material. In the end, 17,388 downloaded images were matched with a street segment, from which 9,252 belonged to detection annotated images. As for detection, we trained and cross-validated with these 9,252 images and tested on the remaining 8,136. 

\subsection{Dataset Statistics}
 
 \begin{figure}
    \subfigure[]{
        \includegraphics[width=0.95\linewidth]{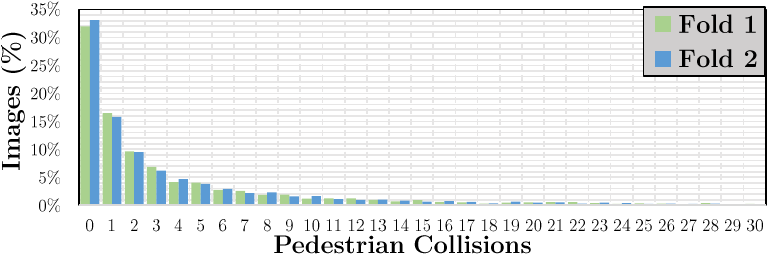}
        \label{fig:accidents_folds}
    }
    \subfigure[]{
        \includegraphics[width=0.95\linewidth]{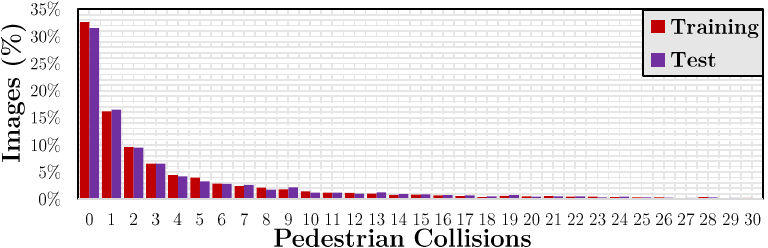}
        \label{fig:accidents_test}
    }
    \caption{\textbf{Pedestrian Collision frequency distribution among training folds (a) and testing set (b).} Figures portray the percentage of images (y-axis) for each amount of pedestrian collisions (x-axis). Our dataset maintains a constant long-tail distribution among the training folds and the testing set. \textbf{Note:} The figures were cut to a maximum of 30 collisions for better visualization.}
    \label{fig:accidents}
\end{figure}

\begin{figure*}[ht!]
    \centering
    \includegraphics[width=0.95\textwidth]{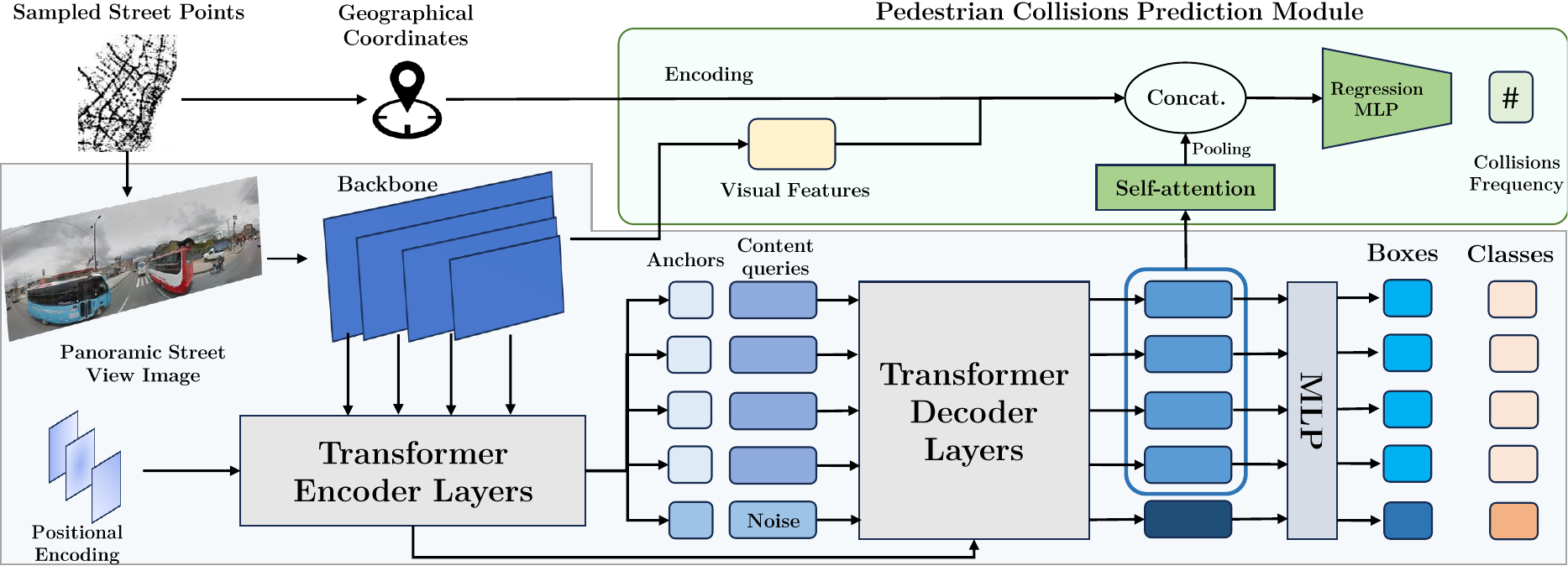}
    \caption{\textbf{STRIDE Baseline Method.} Our model uses DINO \cite{zhang2022dino} (bottom) for object detection on an input panoramic image from a sampled street point. Our pedestrian collision prediction module (top-right) employs self-attention on the output embeddings of DINO's decoder to capture spatial and semantic relationships among object proposals. Additionally, we extract a visual feature from DINO's backbone, and we encode the geographical coordinates of the sampled street point. Finally, we pool the output features of the self-attention layer and concatenate them with the visual features and the encoded coordinates to perform a linear regression and estimate pedestrian collision frequency.}
    \label{fig:overview}
\end{figure*}

Figure \ref{fig:annotations_distr} presents the distribution of the number of annotated bounding boxes per image. We annotated a total of 557,115 objects. Our images contain an average of 56.5 annotations per image, a minimum of 2 boxes, and a maximum of 275 boxes per image. This diverse range of annotations per image allows for a robust representation of various street object distribution in different urban scenes. Furthermore, Figure \ref{fig:classes_distr} shows each semantic class's frequency distribution. We observe a considerable imbalance in the frequency distribution of our classes, as some parts of the street infrastructure, like curbs, street lights, and trees, are naturally more frequent than roundabouts, Bus Rapid Transit (BRT) Stations, and Median barriers. Hence, this long tail distribution is highly representative of real-world-urban scenes.
 
Additionally, our annotated bounding boxes exhibit highly varying sizes. Figure \ref{fig:rel_area} portrays the distribution of the bounding box areas relative to the image size. The absolute areas of our annotated boxes range from 100 to $57.15e^6$ pixels with an average size of $4.37e^5$. Based on the MS COCO standards \cite{mscoco}, the absolute areas of our annotations correspond mostly to large objects (see the Supplementary Material for more detail on box areas distribution according to MS COCO standards). However, due to the considerably large size of our images, most boxes have a low relative area with an average relative area of 0.82\%. This characteristic makes our detection benchmark extremely challenging.

Finally, Figure \ref{fig:accidents_folds} exhibits the histogram with the distribution of the number of pedestrian collisions per image in the training set folds. Figure \ref{fig:accidents_test} exhibits the distribution in the test set. The number of pedestrian collisions corresponding to each image presents a considerable imbalance and an evident long tail in both sets as, naturally, most crossing points have a low pedestrian collision incidence. Our images have a minimum of 0 pedestrian collisions, a maximum of 193 pedestrian collisions, an average of 6.65 pedestrian collisions, and a standard deviation of 14.3. This imbalance in the pedestrian collisions distribution presents another challenge in our data.
%, as it can bias the training and evaluation of accident estimation models.

\subsection{Evaluation Metrics}
% This choice was motivated by the robustness and wide acceptance of the AP metric as the standard for object detection evaluation \cite{zhang2022dino, Cityscapes}.

We use the standard Average Precision (AP) metric from MS COCO \cite{mscoco} to evaluate the object detection task. For the pedestrian collisions prediction task, we adopt the root mean squared error (RMSE) as the primary evaluation metric as it has an increased sensitivity for large errors. Similarly, we avoid using the mean absolute error (MAE) metric since our unbalanced data easily biases it. As an alternative, we propose the Weighted Mean Absolute Error (WMAE) as a more stringent metric that severely penalizes underestimations, thereby addressing the specific challenges posed by our benchmark. We define WMAE as follows:

\begin{equation}
WMAE(y,\hat{y}) = \frac{\sum_{i}^{N}{(y_i+1)|y_i-\hat{y}_i|}}{\sum_{i}^{N}{(y_i+1)}}
\end{equation}

Where $y_i$ is the ground truth value of the $i$th image, $\hat{y}_i$ is the predicted value of the $i$th image, and $N$ is the total number of images.

% ------------------- METHOD ------------------------------
\section{STRIDE Baseline}\label{sec:method}
We propose a multi-task model capable of simultaneously detecting essential street infrastructure objects in panoramic street images and predicting the frequency of pedestrian collisions. Figure \ref{fig:overview} depicts our model. The general intuition behind our method lies in leveraging the visual information captured in panoramic street images to identify built environment elements and their correlation with pedestrian collision occurrences. Thus, we employ a two-stage process; we first utilize the DINO~\cite{zhang2022dino} model to detect various street infrastructure objects in the images. Subsequently, we introduce a pedestrian collision prediction module that exploits self-attention to capture spatial and semantic relationships among the detected objects. By combining the extracted features from both stages, we aim to enhance the accuracy of pedestrian collision prediction.

\subsection{Object Detector}

We build upon DINO~\cite{zhang2022dino}, a state-of-the-art efficient object detection model. As a DETR-like architecture~\cite{carion2020detr}, DINO utilizes a set-prediction approach to predict a fixed-size set $z$ of $\mathcal{N}$ class probability-bounding box pairs using $\mathcal{N}$ input learnable queries. DINO combines multiple enhancements to the original DETR model~\cite{liu2022dabdetr,li2022dn,zhu2021deformable}. First, it includes a mixed query selection approach that leverages information from the backbone and the encoder to initialize anchor boxes as positional embeddings. Additionally, a contrastive denoising training introduces negative and positive noise to ground truth boxes, facilitating bipartite matching and faster convergence. Finally, DINO also integrates box refinement and deformable attention from Deformable-DETR~\cite{zhu2021deformable} to boost efficiency and performance. For further details on the DINO architecture, we refer readers to the original DINO paper~\cite{zhang2022dino}. We adapt and optimize this architecture to detect and classify annotated built environment objects in our panoramic street images.

% For the object detection task, we build upon DINO \cite{zhang2022dino}, an efficient transformer-based model for end-to-end object detection. As a DETR-like architecture, DINO employs a set-prediction approach, which aims to predict a fixed-size set $z$ of $\mathcal{N}$ class probability-bounding box pairs by using $\mathcal{N}$ input learnable queries. DINO surpasses traditional object detectors by incorporating multiple previous enhancements of the DETR architecture \cite{liu2022dabdetr,li2022dn,zhu2021deformable}. First, it introduces a mixed query selection approach that leverages information from the backbone and the encoder to select and initialize anchor boxes as positional embeddings. Additionally, a contrastive denoising training is employed, which introduces negative and positive noise to ground truth boxes and a denoising loss to ease bipartite matching and accelerate convergence. Furthermore, DINO incorporates box refinement and deformable attention from the Deformable-DETR \cite{zhu2021deformable} model to improve efficiency and performance. We refer the reader to the original DINO paper for any further detail on the DINO architecture \cite{zhang2022dino}. We adapt and optimize this architecture to detect and classify the annotated built environment objects in panoramic street images.

\subsection{Pedestrian Collisions Prediction}

We extend DINO with an additional pedestrian collisions prediction module (PCPM) as shown in figure \ref{fig:overview}. This module captures the relationships between the detected objects and the likelihood of pedestrian collisions in a given street image. On the one hand, the PCPM employs a self-attention layer on the output features of DINO's decoder to capture spatial and semantic information. This layer allows the module to focus on relevant regions and learn their interdependencies within identified objects. We exclude the features corresponding to noise queries from the module's processing as we only consider actual object proposals. On the other hand, we extract a visual embedding from the output feature map of DINO's backbone to capture general visual cues from the image. Furthermore, The output features from the self-attention layer are pooled and concatenated with the visual embedding. Additionally, we encode and concatenate the geographical coordinates of the image to include geospatial information. Finally, a regression multi-layer perceptron (MLP) processes the concatenated embeddings and calculates the number of pedestrian collisions associated with the input image. This approach enables us to capture both the global context of the street image and localized details of detected objects to improve the predictive capabilities of our model.

\subsection{Implementation Details}

\noindent{\textbf{Object Detection:}
We train DINO with a ResNet50 backbone and DINO's \textit{4-scale} implementation as we prioritize image resolution over model parameters. To fully exploit the resolution of our images, we train the model using a batch size of 1 and the largest image size allowed by our GPU resources, which is 1,800$\times$5,990 pixels. We trained DINO for 50 epochs in 4 NVIDIA Quadro RTX GPUs with an SGD optimizer, and a learning rate of $1e^{-4}$ decayed by 0.1 after 12 epochs. We use random horizontal flips followed by either a random short side scale augmentation with a 1,600 to 1,800 range, or a short side rescaling between 1,920 and 3,000 pixels with a random crop of 1,800$\times$5,990 pixels. Finally, we rescaled images to 1,800$\times$5,990 for inference.

We discovered that using DINO's pretrained weights on the MS COCO dataset \cite{mscoco} is not beneficial for our specific task due to its substantial differences in class composition with our dataset (each dataset excludes most classes of the other). Hence, we pretrain our model for object detection in Mapillary Vistas \cite{mapillaryvistas} using only the categories with certain similarities to ours. We chose this dataset because it has the best coverage of static street infrastructure categories among previous datasets and provides highly diverse images that include some Latin American scenes.\\

% We provide a detailed list of the chosen categories from Mapillary Vistas in the supplementary material. 

\noindent{\textbf{Pedestrian Collisions Prediction Module:}
We train our module for 20 epochs with a batch size of 5 on 4 Quadro RTX GPUs. We rescale images to 1,800$\times$5,990 and use random horizontal flips for training. We use an $L2$ loss function with an SGD optimizer and a learning rate of $1e^{-4}$ decayed by 0.1 after 15 epochs.}

% To counter collision frequency imbalance during training, we introduce a Weighted Mean Square Error (WMSE) loss $\mathcal{L}_{WMSE}$ to strictly penalize errors in images with a high collision frequency. This loss is similar to our WMAE metric and is defined as follows:

% \begin{equation}
%     \mathcal{L}_{WMAE}(y,\hat{y}) = \frac{\sum_{i}^{N}{(y_i+1)\times(y_i-\hat{y}_i)}}{\sum_{i}^{N}{(y_i+1)}}
% \end{equation}

% --------------------- EXPERIMENTS -------------------------
\section{Experiments}\label{sec:experiments}
\textbf{Experimental Setup:} We train and validate DINO with all the 9,900 annotated images in our dataset and report the standard deviation error across folds. Additionally, we modify MS COCO's standards~\cite{mscoco} of object sizes to our images. We define \textit{small objects} as those with $<0.01\%$ relative area, \textit{medium} as $0.01\%$ to $10\%$ relative area, and \textit{large} as $>10\%$ relative area. We use the same cross-validation scheme for pedestrian collisions prediction and we test on the 8,136 images of the test set.

\subsection{Detection}

\begin{table}[]
\centering
\resizebox{\linewidth}{!}{
    \begin{tabular}{c|cc}
    \hline
    \textbf{Model}  & \textbf{DINO \cite{zhang2022dino}} & \textbf{Deformable-DETR \cite{zhu2021deformable}} \\ \hline
    \textbf{$AP$}      & \textbf{32.59} $\pm1.34$ & 30.28 $\pm0.07$ \\
    \textbf{$AP_{50}$} & \textbf{50.53} $\pm1.76$ & 49.25 $\pm0.22$ \\
    \textbf{$AP_{75}$} & \textbf{34.46} $\pm1.43$ & 31.26 $\pm0.15$ \\
    \textbf{$AP_S$}    & \textbf{18.11} $\pm0.63$ & 16.65 $\pm0.29$ \\  
    \textbf{$AP_M$}    & \textbf{45.10} $\pm0.89$ & 42.23 $\pm0.18$ \\
    \textbf{$AP_L$}    & \textbf{56.40} $\pm4.35$ & 54.18 $\pm0.30$ \\ \hline
    Params. & 47M  & 40M \\ \hline
    \end{tabular}}
    \caption{\textbf{Results in STRIDE's object detection task} of DINO \cite{zhang2022dino} compared with Deformable DETR \cite{zhu2021deformable}. The best performances are shown in bold.}
    \label{tab:detection_results}
\end{table}

Table \ref{tab:detection_results} shows the quantitative results of our best detection model. Our best-performing model achieves a considerably lower $AP$ value compared to DINO's performance on generic benchmarks~\cite{zhang2022dino,mscoco}. This performance drop proves the challenging differences between objects' appearances in large panoramic images and standard images, underscoring the need for more specialized detection models tailored to our object detection task in panoramic urban environments. Moreover, our model has the lowest performance on small and medium objects, despite their high frequency in our dataset. This behavior is due to the image downscaling process, which strongly reduces the absolute area of objects, reaching areas of just 50 pixels. We provide qualitative results examples in the Supplementary Material. We note that DINO correctly identifies and locates most objects in our images but frequently produces false positive box predictions.
% This deviance proves the difficulty of our specific object detection benchmark compared to standard object detection benchmarks and reflects the complex nature of built environment object detection in our panoramic urban images. 

Regarding per-class detection performance, we provide detailed results for each class in the Supplementary Material. Our model performs poorly in low frequent categories such as roundabouts and median barriers. However, we also observe that some highly frequent categories, such as curbs or sidewalks, do not yield high detection performances. We attribute this discrepancy to the high intra-class visual appearance variability of these objects, which are often occluded or appear in small sizes at distant parts of the images. 

Finally, we compare our model with Deformable DETR \cite{zhu2021deformable} by adapting and optimizing it for our task using the same backbone and pretraining scheme. Table \ref{tab:detection_results} also summarizes our overall results; detailed results are presented in the Supplementary Material. As expected, DINO outperforms Deformable-DETR in all metrics within fewer training iterations due to its contrastive denoising training. Nevertheless, regardless of the performance difference, we note that both detectors obtain similar value ranges and relative performances for all $AP$ subtypes among both models. These consistencies validate the reliability of our results for object detection in our benchmark.

\subsection{AutoML Regression}

\begin{table}[]
    \resizebox{\linewidth}{!}{
    \begin{tabular}{cc|cc}
    \hline
    \textbf{Predicted measure} & \textbf{Value} & \textbf{RMSE} & \textbf{WMAE} \\ \hline
    \textbf{Mode}   & 0    & 16.93 $\pm0.34$  & 38.11 $\pm1.45$  \\
    \textbf{Median} & 2    & 16.24 $\pm0.36$  & 36.37 $\pm1.50$  \\
    \textbf{Mean}   & 6.70 & 16.32 $\pm1.03$  & 36.11 $\pm3.80$  \\ \hline
    \end{tabular}}
    \caption{\textbf{Control experiment results for the pedestrian collisions prediction task} by constantly predicting the training set's mode, median, and mean to calculate the statistical lower bounds of our benchmark in our training folds.}
    \label{tab:control_exps}
\end{table}

Initially, we conduct control experiments to establish reference points and statistical lower-bound metrics for our pedestrian collision prediction task. Specifically, we calculate our metrics for predicting statistical measurements of central tendencies, such as the training set's mean, mode, and median, as constant predictions for the validation process. Table \ref{tab:control_exps} shows these results. Additionally, we perform baseline experiments using a model search with AutoML from Python's H2O library to predict pedestrian collisions. We show these results in Table \ref{tab:automl_exps}. First, we run AutoML solely on the geographical coordinates of our images. Our results demonstrate a considerable correlation between the coordinates and pedestrian collision occurrences, which suggests that the model might learn to discern areas with elevated pedestrian collision frequency within the city. This observation also serves as an additional reference point to understand the relationship of geographical location information in pedestrian injury prediction.

Moreover, we study the impact of incorporating DINO's predictions of the number of instances per category in an image (object counts) for pedestrian collision prediction. We observe a notable increase in performance, indicating that a regression model can effectively leverage object presence information to predict pedestrian collisions. This finding underscores the importance of object recognition inputs, as they provide valuable cues to identify risk factors and assess the likelihood of pedestrian collisions. Similarly, we use ground truth counts and achieve a slight performance increase, therefore proving that the model benefits from improved object recognition. We obtain lower error values by including the geographical information as input, once again proving the impact of geographical location on pedestrian collisions prediction. Regardless of the promising performances of these baseline models, the performance of AutoML is highly limited by the lack of visual or spatial information processing.

\begin{table}[]
\resizebox{\linewidth}{!}{
\begin{tabular}{cc|cc}
\hline
\textbf{Coordinates} & \textbf{Object Counts} & \textbf{RMSE} & \textbf{WMAE} \\ \hline
\checkmark & --           & 15.03 $\pm0.20$  & 30.04 $\pm1.32$ \\
--         & DINO         & 13.66 $\pm0.36$  & 27.73 $\pm1.10$ \\
\checkmark & DINO         & 13.58 $\pm0.36$  & 27.65 $\pm1.25$ \\
--         & Ground Truth & 13.53 $\pm0.26$  & 27.33 $\pm1.00$ \\ 
\checkmark & Ground Truth & \textbf{13.44} $\pm0.26$  & \textbf{26.96} $\pm1.02$ \\ \hline
\end{tabular}}
\caption{\textbf{Results of AutoML experiments on the training folds} using geographical coordinates and the number of instances per class (object counts). We indicate the use (\checkmark) or absence (--) of coordinates and counts and whether the counts are obtained from DINO predictions or ground truth annotations. The best results are shown in bold.}
\label{tab:automl_exps}
\end{table}

\subsection{Pedestrian Collisions Prediction Module}

To evaluate the impact of the detection task on pedestrian collision prediction, we first calculate pedestrian collisions directly from the sole backbone and the input coordinates using an MLP. We use the backbone pretrained in ImageNet \cite{imagenet} and train until convergence. The results of this experiment are shown in Table \ref{tab:collisions_head} as \textit{Backbone Regression}. Using the backbone yields a better performance than AutoML's best model, thus proving the importance of visual information for pedestrian collision prediction as it provides general context features to identify risky environments. 

Additionally, we explore various configurations for the pedestrian collisions predictions module to optimize the performance of our model. Initially, we use a linear layer on the outputs of DINO's decoder; this experiment is portrayed in Table \ref{tab:collisions_head} as \textit{Linear Layer}. This configuration outperforms AutoML's best method and the \textit{Backbone Regression} experiment. This result underscores the potential of object detection embeddings for pedestrian collision prediction. Subsequently, we introduce the self-attention layer (shown in Table \ref{tab:collisions_head} as \textit{Self-Att. Layer}) instead of the linear layer, and we note a significant boost in performance for our model, as it captures relationships and dependencies among the spatial and semantic information encoded in the output embeddings of DINO's decoder. Hence, self-attention enables the model to understand better associations among predicted objects which is essential for pedestrian collision prediction. 

Furthermore, we incorporate the visual embeddings extracted from DINO's backbone into the pedestrian collision prediction module (named \textit{Self-Att. + Visual Embd.}). This modification led to a further increase in performance, especially in $WMAE$. The visual embeddings encapsulate general visual information before object location, providing valuable context cues for pedestrian collision prediction. These results demonstrate the importance of leveraging detection and visual information for accurate pedestrian collision estimation in street scenes. Finally, we explore multiple encoding techniques to incorporate geographical coordinates into the model. These techniques included linear projection or positional encoding. However, we observed no significant performance improvement compared to a simple normalization approach. Our results also demonstrate that the increase in model performance requires a significant computational cost as the number of parameters for each module design (shown in Table \ref{tab:collisions_head}) increases with the inclusion of the self-attention and visual features.

\begin{table}[]
    \resizebox{\linewidth}{!}{
    \begin{tabular}{c|cc|c}
    \hline
    \textbf{Module Design}           & \textbf{RMSE}      & \textbf{WMAE} & \textbf{Params. (M)} \\ \hline
    Backbone Regression            & 13.26 $\pm0.11$    & 27.01 $\pm0.80$ & 0.54 \\
    \hline
    Linear Layer                   & 13.11 $\pm0.49$    & 26.07 $\pm1.68$ & 0.11 \\
    Self-Att. Layer                 & 12.79 $\pm0.38$    & 25.51 $\pm0.78$ & 0.44  \\
    \textbf{Self-Att. + Visual Embd.} & \textbf{12.67} $\pm0.33$    & \textbf{24.73} $\pm1.05$ & 1.08\\ \hline
    \end{tabular}}
    \caption{\textbf{Results of the pedestrian collision prediction module in our training folds.} We compare different design choices of our pedestrian collision prediction module. We present the number of parameters corresponding just to the regression module. The best results are shown in bold.}
    \label{tab:collisions_head}
\end{table}

\subsection{Model Testing}

We directly evaluate our best model for pedestrian collision prediction and compare it with our best AutoML method that takes the object counts predicted by DINO as input. Table \ref{tab:collisions_head_test} presents our results. Both models exhibit similar performance on the test and the cross-validation sets. The relative behavior of both models remains consistent, with the prediction module of DINO consistently outperforming the AutoML model. This observation validates the superiority of the prediction module and reinforces the efficacy of calculating pedestrian collisions directly from the visual, spatial, and semantic embeddings provided by DINO. Our findings confirm the validity of our proposal and underscore the value of leveraging visual information from DINO for accurate pedestrian collision prediction. We provide the distribution of the Exact Error among predictions on the test set in the Supplementary Material. Most of our model predictions achieve low error values with a tendency towards slight overestimations. However, the model fails the most in predicting large pedestrian collision values due to the reduced frequency of these samples.

\begin{table}[]
    \resizebox{\linewidth}{!}{
    \begin{tabular}{c|cc}
    \hline
    \textbf{Model}           & \textbf{RMSE}      & \textbf{WMAE}  \\ \hline
    AutoML                & 13.78 $\pm0.02$    & 28.55 $\pm0.12$ \\
    \textbf{STRIDE Baseline} & \textbf{12.88} $\pm0.01$    & \textbf{23.41} $\pm0.40$ \\ \hline
    \end{tabular}}
    \caption{\textbf{Results of STRIDE's baseline on the test set.} We compare the performance of our best AutoML model with the best pedestrian collision prediction module configuration. The best results are shown in bold. }
    \label{tab:collisions_head_test}
\end{table}

\section{Conclusions}
This paper introduces STRIDE, a novel benchmark to improve environmental awareness in autonomous driving by studying the relationship of built environment elements in pedestrian injury prediction. Our framework introduces a multitask approach to simultaneously detect relevant built-environment features and estimate pedestrian collision frequency. We present a new dataset that geographically associates public records of pedestrian collisions with large-scale panoramic street view images, manually annotated for built environment detection of 27 categories. By presenting multiple challenges representative of real-world situations, our benchmark provides a robust testing ground for image-guided pedestrian collision prediction models. To pave the way for future research, we propose a strong baseline that combines a state-of-the-art object detector with an additional collision prediction module. Our experimental validation demonstrates that our model can leverage our detection annotations by capturing interrelations among built environment features to estimate pedestrian collision frequency. Our benchmark promotes the development of autonomous agents capable of predicting pedestrian collision events solely from visual inputs and GPS coordinates, which holds the potential to enhance situation awareness and real-time active collision prevention. Hence, our work is a stepping stone towards improving security in autonomous driving systems, mitigating potential risks, and ensuring safer transportation. \\

%\section{Acknowledgements}

\noindent{\small{\textbf{Acknowledgements:} Nicolás Ayobi acknowledges the support of the 2022 Uniandes-DeepMind scholarship. This research was supported by the Fogarty International Center of the National Institutes of Health (NIH) under award numbers K01TW011782 and 3K01TW011782-01S1. The content is solely the responsibility of the authors and does not necessarily represent the official views of the NIH.}}

{\small
\bibliographystyle{ieee_fullname}
\bibliography{egbib}
}

\end{document}